\documentclass{article}

\usepackage[preprint, nonatbib]{nips_2018}

\usepackage[utf8]{inputenc} 
\usepackage[T1]{fontenc}    
\usepackage{hyperref}       
\usepackage{url}            
\usepackage{booktabs}       
\usepackage{amsfonts}       
\usepackage{nicefrac}       
\usepackage{microtype}      

\usepackage{amsfonts, amssymb, amsmath, amsthm, wasysym, color}
\usepackage{graphicx, subcaption}

\newcommand{\x}{{\bf x}} 
\newcommand{\y}{{\bf y}} 
\newcommand{\z}{{\bf z}} 
\newcommand{\s}{{\bf s}} 
\newcommand{\E}{\mathbb{E}} 
\newcommand{\KL}{\mathrm{KL}} 

\title{Invariant Representations from \\ Adversarially Censored Autoencoders}

\author{
  Ye Wang,\ \ \ Toshiaki Koike-Akino \\
  Mitsubishi Electric Research Laboratories \\
  Cambridge, MA 02139 \\
  \texttt{\{yewang, koike\}@merl.com} \\
  \And
  Deniz Erdogmus \\
  Dept. of Electrical and Computer Engineering \\
  Northeastern University, 
  Boston, MA 02115 \\
  \texttt{erdogmus@ece.neu.edu} \\
}

\begin{document}

\maketitle

\begin{abstract}
We combine conditional variational autoencoders (VAE) with adversarial censoring in order to learn invariant representations that are disentangled from nuisance/sensitive variations. In this method, an adversarial network attempts to recover the nuisance variable from the representation, which the VAE is trained to prevent. Conditioning the decoder on the nuisance variable enables clean separation of the representation, since they are recombined for model learning and data reconstruction. We show this natural approach is theoretically well-founded with information-theoretic arguments. Experiments demonstrate that this method achieves invariance while preserving model learning performance, and results in visually improved performance for style transfer and generative sampling tasks.

\end{abstract}

\section{Introduction}

We consider the problem of learning data representations that are invariant to nuisance variations and/or sensitive features.
Such representations could be useful for fair/robust classification~\cite{zemel2013-fairrep, louppe2016-pivot, xie2017controllable}, domain adaptation~\cite{tzeng2017-AdvDomAdapt, shen2017-AdvRepLearn}, privacy preservation~\cite{hamm2016-minimax, iwasawa2017privacy}, and style transfer~\cite{mathieu2016disentangling}.
We investigate how this problem can be addressed by extensions of the variational autoencoder (VAE) model introduced by~\cite{kingma2013-VAE},
where a generative model is learned as a pair of neural networks: an encoder that produces a representation $\z$ from data $\x$, and a decoder that reconstructs the data $\x$ from the representation $\z$.

A {\em conditional} VAE~\cite{sohn2015learning} can be trained while conditioned on the nuisance/sensitive variable $\s$ (i.e., the encoder and decoder each have $\s$ as an additional input).
In principle, this should yield an encoder that extracts representations $\z$ that are invariant to $\s$, since the corresponding generative model (decoder) implicitly enforces independence between $\s$ and $\z$.
Intuitively, an efficient encoder should learn to exclude information about $\s$ from $\z$, since $\s$ is already provided directly to the decoder.
However, as we demonstrate in our experiments, invariance is not sufficiently achieved in practice, possibly due to approximations arising from imperfect optimization and parametric models.
The adversarial feature learning approach of~\cite{edwards2015-censor} proposes training an {\em unconditioned} autoencoder along with an adversarial network that attempts to recover a binary sensitive variable $\s$ from the representation $\z$.
However, this approach results in a challenging tradeoff between enforcing invariance and preserving enough information in the representation to allow decoder reconstruction and generative model learning.

Our work proposes and investigates the natural combination of adversarial censoring with a {\em conditional} VAE, while also generalizing to allow categorical (non-binary) or continuous $\s$.
Although an adversary is used to enforce invariance between $\z$ and $\s$, the decoder is given both $\s$ and $\z$ as inputs enabling data reconstruction and model learning.
This approach disentangles the representation $\z$ from the nuisance variations $\s$, while still preserving enough information in $\z$ to recover the data $\x$ when recombined with $\s$.
In Section~\ref{sec:adversarial}, we present a theoretical interpretation for adversarial censoring as reinforcement of the representation invariance that is already implied by the generative model of a conditional VAE.
Our experiments in Section~\ref{sec:experiments} quantitatively and qualitatively show that adversarial censoring of a conditional VAE can achieve representation invariance while limiting degradation of model learning performance.
Further, the performance in style transfer and generative sampling tasks appear visually improved by adversarial censoring (see Figures~\ref{fig:change_comp} and~\ref{fig:sampling_comp}).

\subsection{Further Discussion of Related Work}

The variational fair autoencoder of~\cite{louizos2015-VFAE} extends the conditional VAE by introducing an invariance-enforcing penalty term based on maximum mean discrepancy (MMD). However, this approach is not readily extensible to non-binary or continuous $\s$.

Generative adversarial networks (GAN) and the broader concept of adversarial training were introduced by~\cite{goodfellow2014GAN}.
The work of~\cite{mathieu2016disentangling} also combines adversarial training with VAEs to disentangle nuisance variations $\s$ from the representation $\z$.
However, their approach instead attaches the adversary to the output of the decoder, which requires a more complicated training procedure handling sample triplets and swapping representations, but also incorporates the learned similarity concept of~\cite{larsen2016autoencoding}.
Our approach is much simpler to train since the adversary is attached to the encoder directly enforcing representation invariance.

Addressing the problem of learning fair representations~\cite{zemel2013-fairrep},
further work on adversarial feature learning~\cite{louppe2016-pivot, xie2017controllable, hamm2016-minimax, iwasawa2017privacy, tzeng2017-AdvDomAdapt, shen2017-AdvRepLearn}
have used adversarial training to learn invariant representations tailored to classification tasks
(i.e., in comparison to our work, they replace the decoder with a classifier).
However, note that in~\cite{louppe2016-pivot}, the adversary is instead attached to the output of the classifier.
Besides fairness/robustness, domain adaptation~\cite{tzeng2017-AdvDomAdapt, shen2017-AdvRepLearn} and privacy~\cite{hamm2016-minimax, iwasawa2017privacy} are also addressed.
By considering invariance in the context of a VAE, our approach instead aims to produce general purpose representations and does not require additional class labels.

GANs have also been combined with VAEs in many other ways, although not with the aim of producing invariant representations.
However, the following concepts could be combined in parallel with adversarial censoring.
As mentioned earlier, in~\cite{larsen2016autoencoding}, an adversary attached to the decoder learns a similarity metric to enhance VAE training.
In~\cite{makhzani2015-adversarial, mescheder2017-adversarial}, an adversary is used to approximate the Kullback--Leibler (KL)-divergence in the VAE training objective, allowing for more general encoder architectures and latent representation priors.
Both~\cite{donahue2017adversarial} and~\cite{dumoulin2017adversarially} independently propose a method to train an autoencoder using an adversary that tries to distinguish between pairs of data samples and extracted representations versus synthetic samples and the latent representations from which they were generated.

\section{Formulation}

In Section~\ref{sec:CondVAE}, we review the formulation of conditional VAEs as developed by~\cite{kingma2013-VAE, sohn2015learning}.
Sections~\ref{sec:adversarial} and~\ref{sec:KL-term} propose techniques to enforce invariant representations via adversarial censoring and increasing the KL-divergence regularization.

\begin{figure*}[t!]
\centering
\includegraphics[scale=0.45]{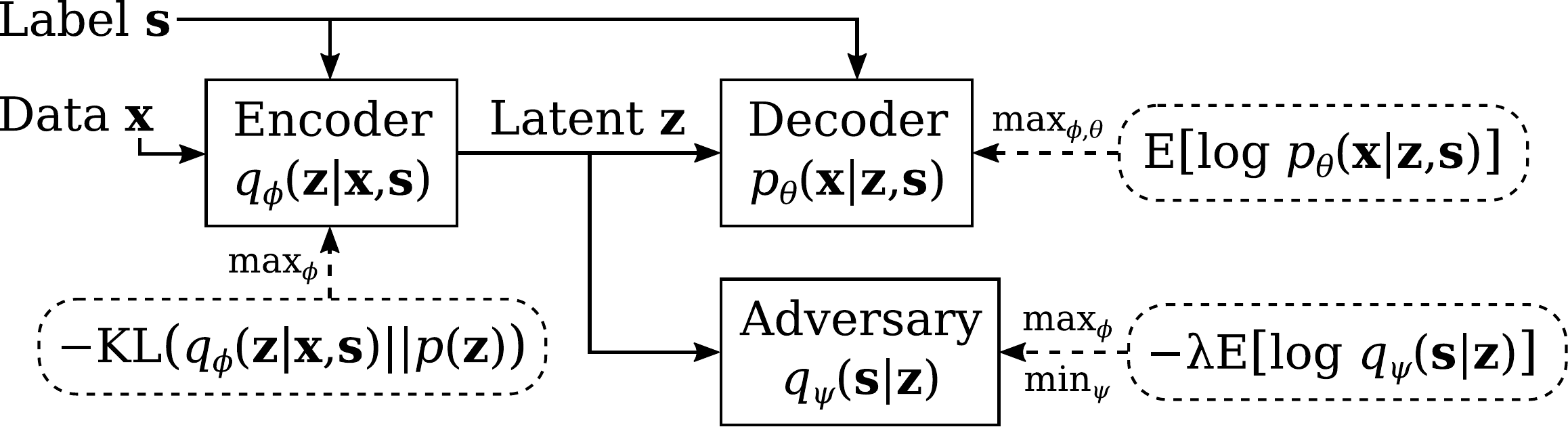}
\caption{Training setup for adversarially censored VAE. The encoder and decoder are trained to maximize the sum of the objective terms (in three dotted boxes), while the adversary is trained to minimize its objective.}
\label{fig:setup}
\end{figure*}

\subsection{Conditional Variational Autoencoders} \label{sec:CondVAE}

The generative model for the data $\x$ involves an observed variable $\s$ and a latent variable $\z$.
The nuisance (or sensitive) variations are modeled by $\s$, while $\z$ captures other remaining information.
Since the aim is to extract a latent representation $\z$ that is free of the nuisance variations in $\s$, these variables are modeled by the joint distribution $(\z, \s, \x) \sim p(\s) p(\z) p_\theta(\x | \z, \s)$, where $\z$ and $\s$ are explicitly made independent.
The generative model $p_\theta(\x | \z, \s)$ is from a parametric family of distributions that is appropriate for the data.
The latent prior $p(\z)$ can be chosen to have a convenient form, such as the standard multivariate normal distribution $\mathcal{N}({\bf 0}, {\bf I})$.
No knowledge or assumptions about the nuisance variable prior $p(\s)$ are needed since it is not directly used in the learning procedure.

The method for learning this model involves maximizing the log-likelihood for a set of training samples $\big\{ (\x_i, \s_i) \big\}^n_{i=1}$ with respect to the conditional distribution
\[
p_\theta(\x | \s) = \int p_\theta(\x | \z, \s) p(\z) d\z.
\]
This objective is analogous to minimizing the KL-divergence between the true conditional distribution $p(\x | \s)$ of the data and the model $p_\theta(\x | \s)$ since
\[
\arg \min_\theta \KL\big( p(\x | \s) \big\| p_\theta(\x | \s) \big) = \arg \max_\theta \E\big[ \log p_\theta(\x | \s) \big],
\]
where the expectation is with respect to $(\x, \s) \sim p(\x | \s) p(\s)$ and can be approximated by $\frac{1}{n}\sum_{i=1}^n \log p_\theta(\x_i | \s_i)$.

Using a variational posterior $q_\phi(\z | \x, \s)$ to approximate the actual posterior $p_\theta(\z | \x, \s) = p(\z) p_\theta(\x | \z, \s) / p_\theta(\x | \s)$, the log-likelihood can be lower bounded by
\begin{align}
\frac{1}{n} \sum_{i=1}^n \log p_\theta(\x_i | \s_i) 
\geq \frac{1}{n} \sum_{i=1}^n \Big[ \E\big[\log p_\theta(\x_i | \s_i, \z_i)\big]
- \KL\big(q_\phi(\z | \x_i, \s_i) \big\| p(\z)\big) \Big] 
=: \mathcal{L}^n(\theta, \phi), \label{eqn:ELBO}
\end{align}
where $\z_i \sim q_\phi(\z | \x_i, \s_i)$ in each expectation.
The quantity $\mathcal{L}^n(\theta, \phi)$ given by~\eqref{eqn:ELBO} is known as the variational or evidence lower bound (ELBO).
The inequality in~\eqref{eqn:ELBO} follows since
\begin{align*}
\frac{1}{n} \sum_{i=1}^n \log p_\theta(\x_i | \s_i) - \mathcal{L}^n(\theta, \phi) 
= \frac{1}{n} \sum_{i=1}^n \KL\big(q_\phi(\z | \x_i, \s_i) \big\| p_\theta(\z | \x_i, \s_i)\big) \geq 0.
\end{align*}
Thus, by optimizing both $p_\theta(\x | \z, \s)$ and $q_\phi(\z | \x, \s)$ to maximize the lower bound $\mathcal{L}^n(\theta, \phi)$, while $p_\theta(\x | \s)$ is trained toward the true conditional distribution $p(\x|\s)$ of the data, $q_\phi(\z | \x, \s)$ is trained toward the corresponding posterior $p_\theta(\z | \x, \s)$.

In the VAE architecture, the generative model (decoder) $p_\theta(\x | \z, \s)$ and variational posterior (encoder) $q_\phi(\z | \x, \s)$ are realized as neural networks that take as input $(\z, \s)$ and $(\x, \s)$, respectively, as illustrated in Figure~\ref{fig:setup}, and output the parameters of their respective distributions.
This architecture is specifically a {\em conditional} VAE, since the encoding and decoding are conditioned on the nuisance variable $\s$. 

When the encoder is realized as conditionally Gaussian:
\begin{equation} \label{eqn:gaussian_encoder}
q_\phi(\z | \x, \s) = \mathcal{N}(\z; \boldsymbol{\mu}_\phi(\x, \s), \mathbf{\Sigma}_\phi(\x, \s)),
\end{equation}
where the mean vector $\boldsymbol{\mu}$ and diagonal covariance matrix $\mathbf{\Sigma}$ are determined as a function of $(\x,\s)$, and the latent variable distribution is set to the standard Gaussian $p(\z) = \mathcal{N}({\bf 0}, {\bf I})$, the KL-divergence term in~\eqref{eqn:ELBO} can be analytically derived and differentiated~\cite{kingma2013-VAE}.
However, the expectations in~\eqref{eqn:ELBO} must be estimated by sampling.

Hence, the learning procedure maximizes a sampled approximation of the ELBO $\mathcal{L}^n(\theta, \phi)$, given by
\begin{align} \label{eqn:basicVAEobj}
\max_{\theta,\phi} & \ \frac{1}{n} \sum_{i=1}^n \mathcal{L}_i(\theta, \phi) \\
\approx \max_{\theta,\phi} \mathcal{L}^n(\theta, \phi) 
& \lessapprox \max_{\theta} \frac{1}{n} \sum_{i=1}^n \log p_\theta(\x_i | \s_i), \nonumber
\end{align}
where, for $i \in \{1, \ldots, n\}$,
\begin{align} \label{eqn:lossterms}
\mathcal{L}_i(\theta, \phi) := - \KL\big(q_\phi(\z | \x_i, \s_i) \big\| p(\z)\big) 
+ \frac{1}{k}\sum_{j=1}^k \log p_\theta(\x_i | \s_i, \z_{i,j}),
\end{align}
which approximates the expectations in~\eqref{eqn:ELBO} by sampling $\{\z_{i,j}\}_{j=1}^k \stackrel{\text{iid}}{\sim} q_\phi(\z | \x_i, \s_i)$.

\subsection{Representation Invariance via Adversarial Censoring} \label{sec:adversarial}

In principle, optimal training with ideal parametric approximations should result in an encoder $q_\phi(\z | \x, \s)$ that accurately approximates the true posterior $p_\theta(\z | \x, \s)$, for which $\z$ and $\s$ are independent by construction.
Thus, the theoretically optimal encoder should produce a representation $\z$ that is independent of the nuisance variable $\s$.
In practice, however, since the encoder is realized as a parametric approximation and globally optimal convergence cannot be guaranteed, we often observe that the representation $\z$ produced by the trained encoder is significantly correlated with the nuisance variable $\s$.
Further, one may wish to train an encoder $q_\phi(\z|\x)$ that does not use $\s$ as an input, to allow the representation $\z$ to be generated from the data $\x$ alone.
However, this additional restriction on the encoder may increase the challenge of extracting invariant representations.

Invariance could be be enforced by minimizing the mutual information $I(\s;\z)$ where $\z \sim q_\phi(\z | \x, \s)$ is the latent representation generated by the encoder.
Mutual information can be subtracted from the lower bound of~\eqref{eqn:ELBO}, yielding
\[
\frac{1}{n} \sum_{i=1}^n \log p_\theta(\x_i | \s_i) \geq \mathcal{L}^n(\theta, \phi) - \frac{1}{n} \sum_{i=1}^n I(\s_i; \z_i),
\]
where equality is still met for $q_\phi(\z | \x, \s) = p_\theta(\z | \x, \s)$.
Thus, incorporating a mutual information penalty term into the lower bound does not, in principle, change the theoretical maximum.
However, since computing mutual information is generally intractable, we apply the approximation technique of~\cite{barber2003-IMalgorithm}, which utilizes a variational posterior $q_\psi(\s|\z)$ and the lower bound
\begin{equation} \label{eqn:MI_estimate}
I(\s; \z) \geq h(\s) + \E\big[\log q_\psi(\s|\z)\big],
\end{equation}
where equality is met for $q_\psi(\s|\z)$ equal to the actual posterior $p(\s|\z)$ for which the expectation and entropies are defined with respect to.
Hence, maximizing $\E\big[\log q_\psi(\s|\z)\big]$ over the variational posterior $q_\psi(\s|\z)$, which can also be similarly realized as a neural network, yields an approximation of $I(\s; \z) - h(\s) = - h(\s | \z)$.
The entropy $h(\s)$, although generally unknown, is constant with respect to the optimization variables.
Incorporating this variational approximation of the mutual information penalty into~\eqref{eqn:basicVAEobj}, modulo dropping the constant $h(\s)$, results in the adversarial training objective
\begin{align} \label{eqn:censorVAEobj}
\max_{\theta,\phi} \min_\psi \frac{1}{n} \sum_{i=1}^n \Big[ \mathcal{L}_i(\theta, \phi) - \frac{\lambda}{k}\sum_{j=1}^k \log q_\psi(\s_i|\z_{i,j}) \Big],
\end{align}
where $\{\z_{i,j}\}_{j=1}^k \stackrel{\text{iid}}{\sim} q_\phi(\z | \x_i, \s_i)$ are the same samples used for $\mathcal{L}_i(\theta, \phi)$ as given by~\eqref{eqn:lossterms}, and $\lambda > 0$ is a parameter that controls the emphasis on invariance.
Note that when $\s$ is a categorical variable (e.g., a class label), the additional, adversarial network to realize the variational posterior $q_\psi(\s|\z)$ is essentially just a classifier trained (by minimizing cross-entropy loss) to recover $\s$ from the representation $\z$ generated by the encoder.
In this approach, the VAE is adversarially trained to maximize the cross-entropy loss of this classifier combined with the original objective given by~\eqref{eqn:basicVAEobj}.
Figure~\ref{fig:setup} illustrates the overall VAE training framework including adversarial censoring.

\subsection{Invariance via KL-divergence Censoring} \label{sec:KL-term}

Another approach to enforce invariance is to introduce a hyperparameter $\gamma > 1$ to increase the weight of the KL-divergence terms in~\eqref{eqn:lossterms}, yielding the alternative objective terms
\begin{align} \label{eqn:KL-lossterms}
\mathcal{L}_i^{\gamma}(\theta, \phi) := - \gamma \ \KL\big(q_\phi(\z | \x_i, \s_i) \big\| p(\z)\big) + \frac{1}{k}\sum_{j=1}^k \log p_\theta(\x_i | \s_i, \z_{i,j}),
\end{align}
for which~\eqref{eqn:lossterms} is the special case when $\gamma = 1$.
The KL-divergence terms $\KL\big(q_\phi(\z | \x_i, \s_i) \big\| p(\z)\big)$ can be interpreted as regularizing the variational posterior toward the latent prior, which encourages the encoder to generate representations $\z$ that are invariant to not only $\s$ but also the data $\x$.
While increasing $\gamma$ further encourages invariant representations, it potentially disrupts model learning, since the overall dependence on the data is affected.

\section{Experiments} \label{sec:experiments}

\begin{figure}[ht!]
\centering
\begin{subfigure}[t]{0.49\textwidth}
    \centering
    \includegraphics[scale=0.45]{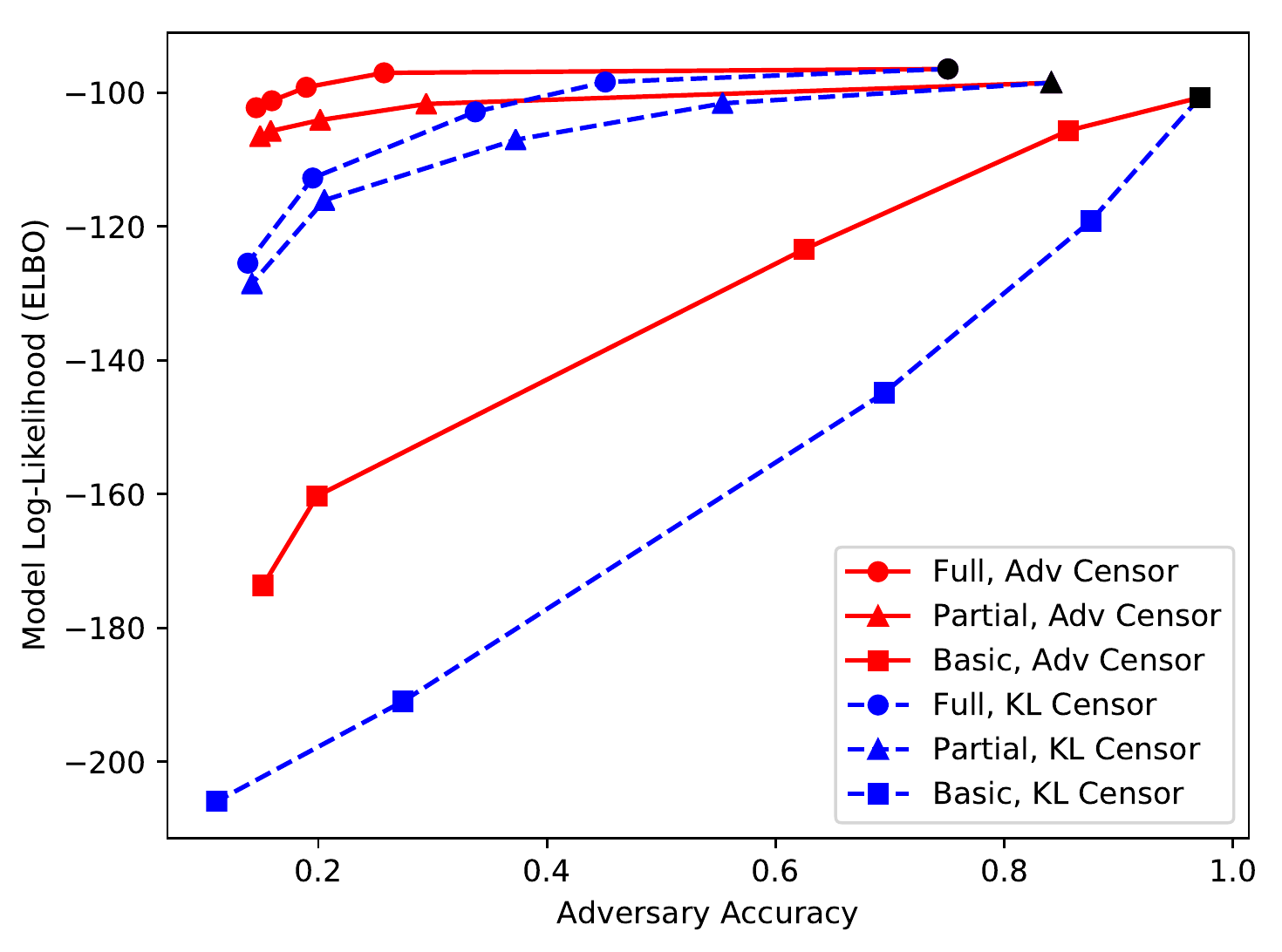}
    \caption{Adversary Accuracy vs ELBO}
\end{subfigure}
\begin{subfigure}[t]{0.49\textwidth}
    \centering
    \includegraphics[scale=0.45]{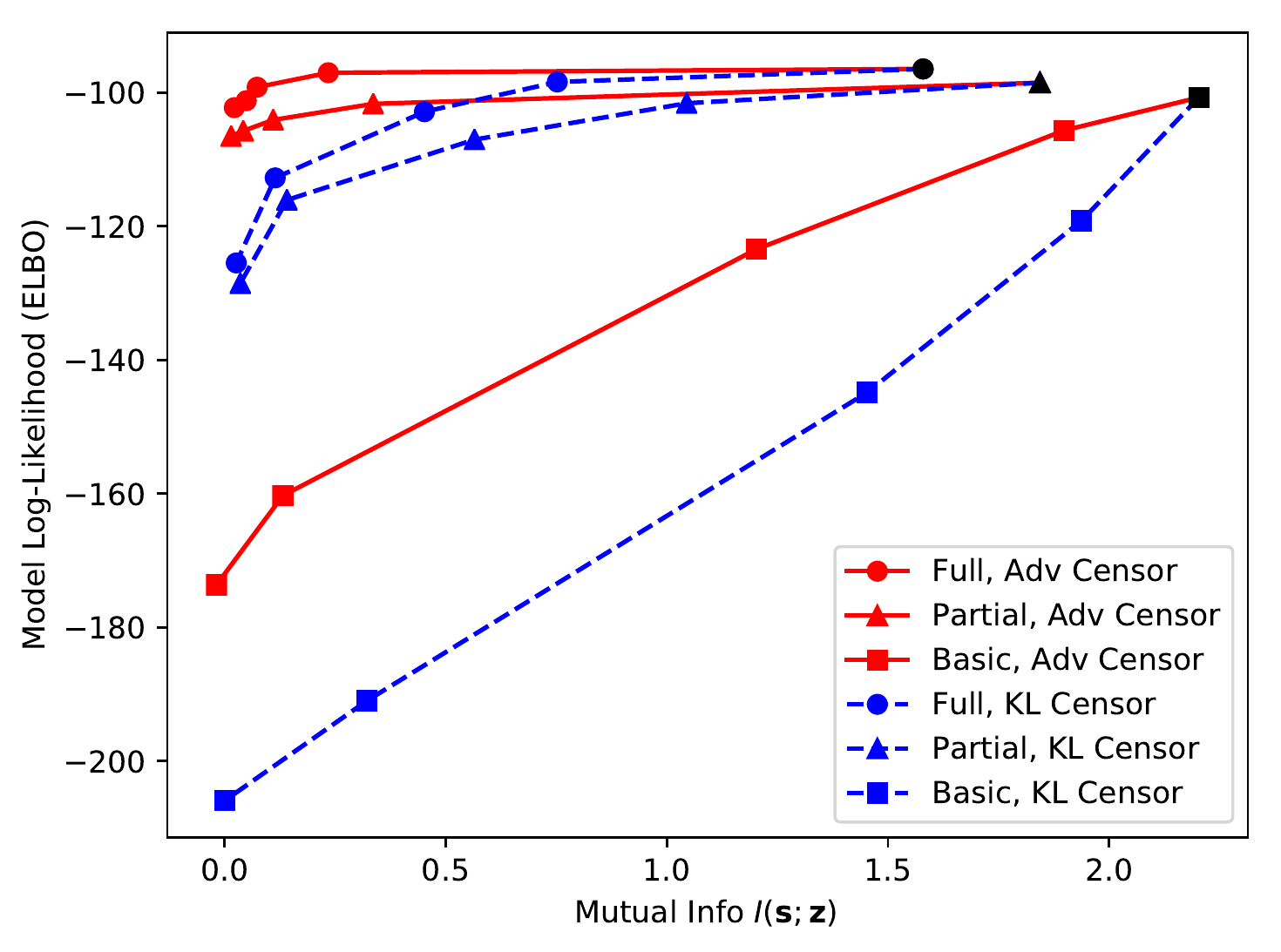}
    \caption{Mutual Information vs ELBO}
\end{subfigure}
\caption{Quantitative performance comparison. Smaller values along the x-axes correspond to better invariance. Larger values along the y-axis (ELBO) correspond to better model learning.}
\label{fig:tradeoffs}
\end{figure}

We evaluate the performance of various VAEs for learning invariant representations,
under several scenarios for conditioning the encoder and/or decoder on the sensitive/nuisance variable $\s$:
\begin{itemize}
\item {\bf Full:} Both the encoder and decoder are conditioned on $\s$.
In this case, the decoder is the generative model $p_\theta(\x | \z, \s)$
and the encoder is the variational posterior $q_\phi(\z | \x, \s)$
as described in Section~\ref{sec:CondVAE}.
\item {\bf Partial:} Only the decoder is conditioned on $\s$.
This case is similar to the previous, except that the encoder approximates the variational posterior $q_\phi(\z | \x)$ without $\s$ as an input.
\item {\bf Basic (unconditioned):} Neither the encoder nor decoder are conditioned on $\s$.
This baseline case is the standard, unconditioned VAE where $\s$ is not used as an input.
\end{itemize}

In combination with these VAE scenarios, we also examine several approaches for encouraging invariant representations:
\begin{itemize}
\item {\bf Adversarial Censoring:} This approach, as described in Section~\ref{sec:adversarial}, introduces an additional network that attempts to recover $\s$ from the representation $\z$.
The VAE and this additional network are adversarially trained according to the objective given by~\eqref{eqn:censorVAEobj}.
\item {\bf KL Censoring:} This approach, as described in Section~\ref{sec:KL-term}, increases the weight on the KL-divergence terms, using the alternative objective terms given by~\eqref{eqn:KL-lossterms}.
\item {\bf Baseline (none):} As a baseline, the VAE is trained according to the original objective given by~\eqref{eqn:basicVAEobj} without any additional modifications to enforce invariance.
\end{itemize}

\begin{figure}[ht!]
\centering
\begin{subfigure}[c]{0.325\textwidth}
    \centering
    \includegraphics[scale=0.365]{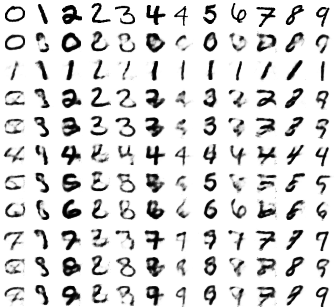}
    \caption{Partial -- Baseline}\label{fig:partbase_change}
\end{subfigure}
\begin{subfigure}[c]{0.325\textwidth}
    \centering
    \includegraphics[scale=0.365]{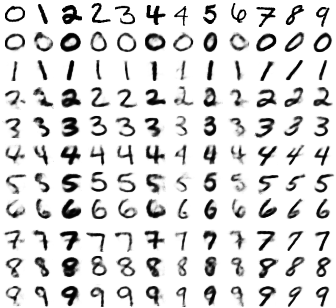}
    \caption{Partial -- Adversarial censoring}\label{fig:partcen_change}
\end{subfigure}
\begin{subfigure}[c]{0.325\textwidth}
    \centering
    \includegraphics[scale=0.365]{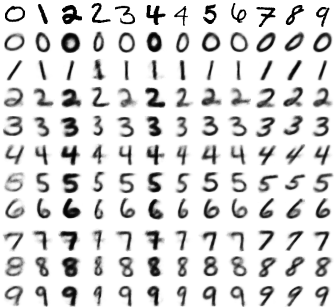}
    \caption{Partial -- KL censoring}\label{fig:partKL_change}
\end{subfigure}
\vspace{1em} 

\begin{subfigure}[c]{0.325\textwidth}
    \centering
    \includegraphics[scale=0.365]{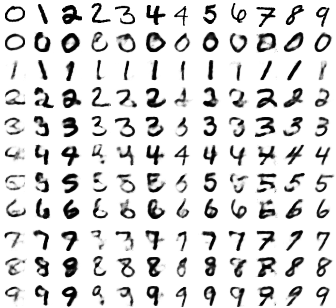}
    \caption{Full -- Baseline}\label{fig:fullbase_change}
\end{subfigure}
\begin{subfigure}[c]{0.325\textwidth}
    \centering
    \includegraphics[scale=0.365]{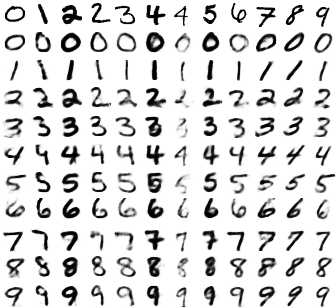}
    \caption{Full -- Adversarial censoring}\label{fig:fullcen_change}
\end{subfigure}
\begin{subfigure}[c]{0.325\textwidth}
    \centering
    \includegraphics[scale=0.365]{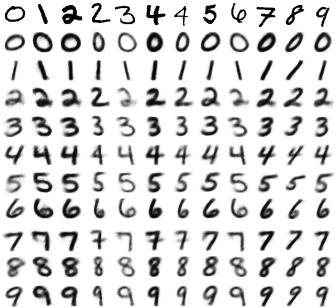}
    \caption{Full -- KL censoring}\label{fig:fullKL_change}
\end{subfigure}
\caption{Style transfer with conditional VAEs.
The top row within each image shows the original test set examples input the encoder, while the other rows show the corresponding output of the decoder when conditioned on different digit classes $\{0, \ldots, 9\}$.
} \label{fig:change_comp}
\end{figure}

\subsection{Dataset and Network Details}

We use the MNIST dataset, which consists of 70,000 grayscale, $28 \times 28$ pixel images of handwritten digits and corresponding labels in $\{0, \ldots 9\}$.
We treat the vectorized images in $[0,1]^{784}$ as the data $\x$, while the digit labels serve as the nuisance variable $\s$.
Thus, our objective is to train VAE models that learn representations $\z$ that capture features (i.e., handwriting style) invariant of the digit class $\s$.

We use basic, multilayer perceptron architectures to realize the VAE (similar to the architecture used in~\cite{kingma2013-VAE}) and the adversarial network.
This allows us to illustrate how the performance of even very simple VAE architectures can be improved with adversarial censoring.
We choose the latent representation $\z$ to have 20 dimensions, with its prior set as the standard Gaussian, i.e., $p(\z) = \mathcal{N}({\bf 0}, {\bf I})$.
The encoder, decoder, and adversarial networks each use a single hidden layer of 500 nodes with the $\tanh$ activation function.
In the scenarios where the encoder (or decoder) is conditioned on the nuisance variable, the one-hot encoding of $\s$ is concatenated with $\x$ (or $\z$, respectively) to form the input.
The adversarial network uses a 10-dimensional softmax output layer to produce the variational posterior $q_\psi(\s|\z)$.

We use the encoder to realize the conditionally Gaussian variational posterior given by~\eqref{eqn:gaussian_encoder}.
The encoder network produces a 40-dimensional vector (with no activation function applied) that represents the mean vector $\boldsymbol{\mu}$ concatenated with the log of the diagonal of the covariance matrix $\mathbf{\Sigma}$.
This allows us to compute the KL-divergence terms in~\eqref{eqn:lossterms} analytically as given by~\cite{kingma2013-VAE}.

The output layer of the decoder network has 784 nodes and applies the sigmoid activation function, matching the size and scale of the images.
We treat the decoder output, denoted by $\y = f_\theta(\s, \z)$, as parameters of a generative model $p_\theta(\x | \s, \z)$ given by
\[
\log p_\theta(\x | \s, \z) = \sum_{i=1}^{784} x_i \log y_i + (1-x_i) \log (1-y_i),
\]
where $x_i$ and $y_i$ are the components of $\x$ and $\y$, respectively.
Although not strictly binary, the MNIST images are nearly black and white, allowing this Bernoulli generative model to be a reasonable approximation.
We directly display $\y$ to generate the example output images.

We implemented these experiments with the Chainer deep learning framework~\cite{chainer}.
The networks were trained over the 60,000 image training set for 100 epochs with 100 images per batch, while evaluation and example generation were performed with the 10,000 image test set.
The adversarial and VAE networks were each updated alternatingly once per batch with Adam~\cite{kingma2014adam}.
Relying on stochastic estimation over each batch, we set the sampling parameter $k = 1$ in~\eqref{eqn:lossterms},~\eqref{eqn:censorVAEobj}, and~\eqref{eqn:KL-lossterms}.

\begin{figure}[ht!]
\centering
\begin{subfigure}[c]{0.325\textwidth}
    \centering
    \includegraphics[scale=0.42]{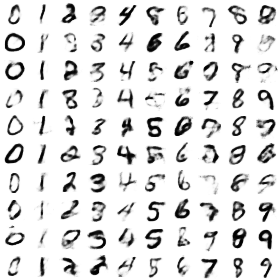}
    \caption{Partial -- Baseline}\label{fig:partbase_sampled}
\end{subfigure}
\begin{subfigure}[c]{0.325\textwidth}
    \centering
    \includegraphics[scale=0.42]{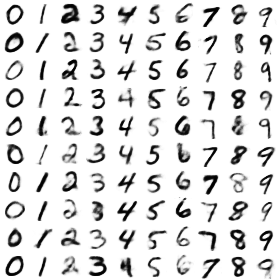}
    \caption{Partial -- Adversarial censoring}\label{fig:partcen_sampled}
\end{subfigure}
\begin{subfigure}[c]{0.325\textwidth}
    \centering
    \includegraphics[scale=0.42]{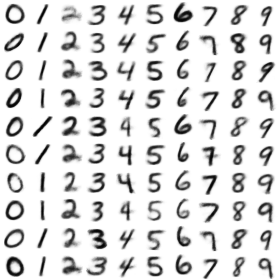}
    \caption{Partial -- KL censoring}\label{fig:partKL_sampled}
\end{subfigure}
\vspace{1em} 

\begin{subfigure}[c]{0.325\textwidth}
    \centering
    \includegraphics[scale=0.42]{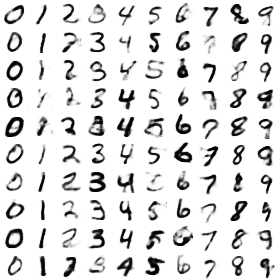}
    \caption{Full -- Baseline}\label{fig:fullbase_sampled}
\end{subfigure}
\begin{subfigure}[c]{0.325\textwidth}
    \centering
    \includegraphics[scale=0.42]{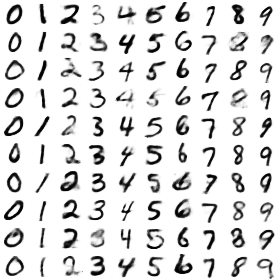}
    \caption{Full -- Adversarial censoring}\label{fig:fullcen_sampled}
\end{subfigure}
\begin{subfigure}[c]{0.325\textwidth}
    \centering
    \includegraphics[scale=0.42]{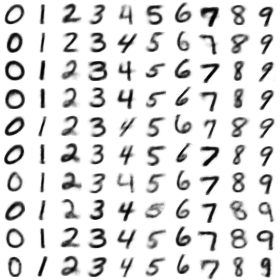}
    \caption{Full -- KL censoring}\label{fig:fullKL_sampled}
\end{subfigure}
\caption{Generative sampling with conditional VAEs.
Latent representations $\z$ are sampled from $p(\z) = \mathcal{N}({\bf 0}, {\bf I})$ and input to the decoder to generate synthetic images, with the decoder conditioned on selected digit classes in $\{0, \ldots, 9\}$.
} \label{fig:sampling_comp}
\end{figure}

\begin{figure}[ht!]
\centering
\begin{subfigure}[c]{0.325\textwidth}
    \centering
    \includegraphics[scale=0.42]{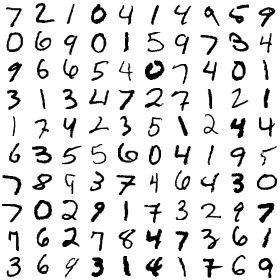}
    \caption{MNIST Examples}\label{fig:mnist_examples}
\end{subfigure}
\begin{subfigure}[c]{0.325\textwidth}
    \centering
    \includegraphics[scale=0.42]{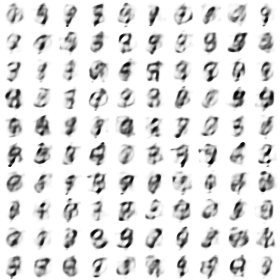}
    \caption{Basic -- Adversarial censoring}\label{fig:nonecen_sampled1}
\end{subfigure}
\begin{subfigure}[c]{0.325\textwidth}
    \centering
    \includegraphics[scale=0.42]{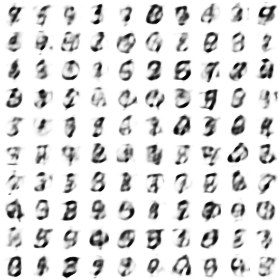}
    \caption{Basic -- Adversarial censoring}\label{fig:nonecen_sampled2}
\end{subfigure}
\vspace{1em} 

\begin{subfigure}[c]{0.325\textwidth}
    \centering
    \includegraphics[scale=0.42]{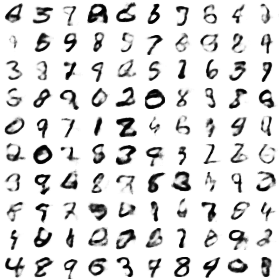}
    \caption{Basic -- Baseline}\label{fig:nonebase_sampled}
\end{subfigure}
\begin{subfigure}[c]{0.325\textwidth}
    \centering
    \includegraphics[scale=0.42]{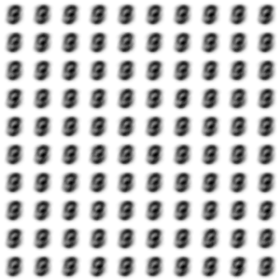}
    \caption{Basic -- KL censoring}\label{fig:noneKL_sampled1}
\end{subfigure}
\begin{subfigure}[c]{0.325\textwidth}
    \centering
    \includegraphics[scale=0.42]{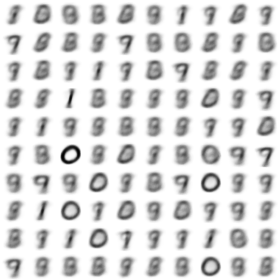}
    \caption{Basic -- KL censoring}\label{fig:noneKL_sampled2}
\end{subfigure}
\caption{Generative sampling with {\em unconditioned} (``basic'') VAEs.
Attempting to censor an unconditioned VAE results in severely degraded model performance.
} \label{fig:sampling_uncond}
\end{figure}

\subsection{Evaluation Methods}

We quantitatively evaluate the trained VAEs for how well they:
\begin{itemize}
\item {\bf Learn the data model:} We measure this with the ELBO score estimated by computing $\frac{1}{n}\sum_{i=1}^n \mathcal{L}_i(\theta, \phi)$ over the test data set (see~\eqref{eqn:basicVAEobj} and~\eqref{eqn:lossterms}).
\item {\bf Produce invariant representations:} We measure this via the adversarial approach described in Section~\ref{sec:adversarial}.
Even when {\em not} using adversarial censoring, we still train an adversarial network in parallel (i.e., its loss gradients are {\em not} fed back into the main VAE training) that attempts to recover the sensitive variable $\s$ from the representation $\z$.
The classification accuracy and cross-entropy loss of the adversarial network provide measures of invariance.
Since the digit class $\s$ is uniformly distributed over $\{0, \ldots, 9\}$, the entropy $h(\s)$ is equal to $\log(10)$ and can be combined with the cross-entropy loss (see~\eqref{eqn:MI_estimate} and~\cite{barber2003-IMalgorithm}) to yield an estimate of the mutual information $I(\s; \z)$, which we report instead.
\end{itemize}

The VAEs are also qualitatively evaluated with the following visual tasks:
\begin{itemize}
\item {\bf Style Transfer (Digit Change):} An image $\x$ from the test set is input to the encoder to produce a representation $\z$ by sampling from $q_\phi(\z | \x, \s)$.
Then, the decoder is applied to produce the image $\y = f_\theta(\s', \z)$, while {\em changing} the digit class to $\s' \in \{0, \ldots, 9\}$.
\item {\bf Generative Model Sampling:} A synthetic image is generated by first sampling a latent variable $\z$ from the prior $p(\z) = \mathcal{N}({\bf 0}, {\bf I})$, and then applying the decoder to produce the image $\y = f_\theta(\s, \z)$ for a selected digit class $\s \in \{0, \ldots, 9\}$.
\end{itemize}

\subsection{Results and Discussion}

Figure~\ref{fig:tradeoffs} presents the quantitative performance comparison for the various combinations of VAEs with full ($\CIRCLE$), partial ($\blacktriangle$), or no conditioning ({\small$\blacksquare$}), and with invariance encouraged by adversarial censoring ({\color{red} red ---}), KL censoring ({\color{blue} blue -{}-{}-}), or nothing (black).
Each pair of red and blue curves represent varying emphasis on enforcing invariance (as the parameters $\lambda$ and $\gamma$ are respectively changed) and meet at a black point corresponding to the baseline (no censoring) case (where $\lambda = 0$ and $\gamma = 1$).

Unsurprisingly, the baseline, unconditioned VAE produces a representation $\z$ that readily reveals the digit class $\s$ ($97.1\%$ accuracy), since otherwise image reconstruction by the decoder would be difficult.
However, even when partially or fully conditioned on $\s$, the baseline VAEs still significantly reveal $\s$ (partial: $84.1\%$, full: $75.1\%$ accuracies).
Both adversarial and KL censoring are effective at enforcing invariance, with adversarial accuracy approaching chance and mutual information approaching zero as the parameters $\lambda$ and $\gamma$ are respectively increased.
However, the adversarial approach has less of an impact on the model learning performance (as measured by the ELBO score).
With conditional VAEs, adversarial censoring achieves invariance while having only a small impact on the ELBO score,
and appears to visually improve performance (particularly for the partially conditioned case) in the style transfer and sampling tasks as shown in Figures~\ref{fig:change_comp} and~\ref{fig:sampling_comp}.
The worse model learning performance with KL censoring seems to result in blurrier (although seemingly cleaner) images, as also shown in Figures~\ref{fig:change_comp} and~\ref{fig:sampling_comp}.
Attempting to censor a basic (unconditioned) autoencoder (as proposed by~\cite{edwards2015-censor}) rapidly degrades model learning performance, which manifests as severely degraded sampling performance as shown in Figure~\ref{fig:sampling_uncond}.

The results in Figures~\ref{fig:change_comp} and~\ref{fig:sampling_comp} correspond to specific points in Figure~\ref{fig:tradeoffs} as follows: (a) baseline $\blacktriangle$, (b) left-most {\color{red} ---\!\!$\blacktriangle$\!\!---} ($\lambda = 20$), (c) left-most {\color{blue} -{}-\!$\blacktriangle$\!-{}-} ($\gamma = 8$), (d) baseline $\CIRCLE$, (e) left-most {\color{red} ---\!\!$\CIRCLE$\!\!---} ($\lambda = 20$), (f) left-most {\color{blue} -{}-$\CIRCLE$-{}-} ($\gamma = 8$).
Figure~\ref{fig:sampling_uncond} results correspond to points in Figure~\ref{fig:tradeoffs} as follows: (a) MNIST test examples, (b-c) two left-most {\color{red} ---\!\!{\small$\blacksquare$}\!\!---} ($\lambda = 100, 50$), (d) baseline {\small$\blacksquare$}, (e-f) two left-most {\color{blue} -{}-{\small$\blacksquare$}-{}-} ($\gamma = 50, 20$).
Note that larger values for the $\lambda$ and $\gamma$ parameters were required for the unconditioned VAEs to achieve similar levels of invariance as the conditioned cases.

\section{Conclusion}
The natural combination of conditional VAEs with adversarial censoring is a theoretically well-founded method to generate invariant representations that are disentangled from nuisance variations.
Conditioning the decoder on the nuisance variable $\s$ allows the representation $\z$ to be cleanly separated and model learning performance to be preserved, since $\s$ and $\z$ are both used to reconstruct the data $\x$.
Training VAEs with adversarial censoring visually improved performance in style transfer and generative sampling tasks.


\bibliography{refs}
\bibliographystyle{abbrv}

\end{document}